\pdfoutput=1

\documentclass[11pt]{article}

\usepackage{acl}

\usepackage{times}
\usepackage{latexsym}

\usepackage[T1]{fontenc}

\usepackage[utf8]{inputenc}

\usepackage{microtype}

\usepackage{inconsolata}
\usepackage{url}
\usepackage{color}
\usepackage{xcolor}
\usepackage{booktabs}
\usepackage{graphicx}
\usepackage{multirow}
\usepackage{multicol}
\usepackage{listings}
\usepackage{array}
\usepackage[font+=small]{subcaption}
\usepackage{tabularx} 
\usepackage{amsfonts}
\usepackage{amsmath}
\usepackage{amssymb}
\usepackage{caption}
\usepackage{algorithm}
\usepackage{algorithmicx}
\usepackage{algpseudocode}
\usepackage{xparse}
\usepackage{tcolorbox}
\usepackage{tikz}
\tcbuselibrary{skins}
\usepackage[graphicx]{realboxes}
\usepackage{adjustbox}
\usepackage{wrapfig}
\tcbuselibrary{raster}
\usepackage{pifont}
\usepackage{xcolor}         
\usepackage[normalem]{ulem}
\usepackage{CJKutf8}
\usepackage{amsthm}

\theoremstyle{definition}
\newtheorem{definition}{Definition}[section]

\theoremstyle{remark}
\newtheorem*{remark}{Remark}

\newif\ifeqsmall
\eqsmallfalse
\newif\iftabsmall
\tabsmallfalse
\eqsmalltrue

\definecolor{darkpastelgreen}{rgb}{0.01, 0.75, 0.24}
\newif\ifcomment
\commenttrue
\definecolor{Brick}{HTML}{B73239}
\definecolor{Cobalt}{HTML}{60007B}
\NewDocumentCommand{\heng}{ mO{} }{\ifcomment\textcolor{red}{\textsuperscript{\textit{Heng}}\textsf{\textbf{\small[#1]}}}\fi}

\NewDocumentCommand{\pengfei}{ mO{} }{\ifcomment\textcolor{Cobalt}{\textsuperscript{\textit{Pengfei}}\textsf{\textbf{\small[#1]}}}\fi}
\NewDocumentCommand{\problem}{}{self information updating}

\NewDocumentCommand{\ProBlem}{}{Self Information Updating}
\NewDocumentCommand{\shortproblem}{}{SIU}

\title{Information Association for Language Model Updating by Mitigating LM-Logical Discrepancy}

\author{Pengfei Yu \\
  University of Illinois Urbana-Champaign\\
  \texttt{pengfei4@illinois.edu} \\\\\And
  Heng Ji \\
  University of Illinois Urbana-Champaign\\
  \texttt{hengji@illinois.edu} \\}

\begin{document}
\maketitle
\begin{abstract}
Large Language Models~(LLMs) struggle with providing current information due to the outdated pre-training data. Existing methods for updating LLMs, such as knowledge editing and continual fine-tuning, have significant drawbacks in generalizability of new information and the requirements on structured updating corpus. We identify the core challenge behind these drawbacks: the LM-logical discrepancy featuring the difference between language modeling probabilities and logical probabilities. 
To evaluate and address the core challenge, we propose a new task formulation of the information updating task that only requires the provision of an unstructured updating corpus and evaluates the performance of information updating on the generalizability to question-answer pairs pertaining to the updating information.
We further propose a novel and effective pipeline approach for the task, highlighting a self-prompting-based question-answer generation process and a associative distillation methods to bridge the LM-logical discrepancy.
We develop two datasets for evaluation, one sourced from news articles published in March and April 2023\footnote{the latest available news by the time of dataset collection}, and the other from the Natural Questions benchmark.
Experimental results demonstrate the superiority of our approach, significantly increasing the factual consistency score (on a scale from 0 to 1) by up to 0.16. Furthermore, our method effectively mitigates forgetting utilizing a compact replay buffer with only 2.3\% of the training tokens.
\end{abstract}

\section{Introduction}
\begin{table}[!ht]
  \iftabsmall\small\fi
  \centering
  \begin{tabular}{p{0.95\linewidth}}
    \toprule
    \textit{New Information}: Louisville Metro Police Department Officer Nickolas Wilt is \textbf{in critical condition after undergoing brain surgery} following a shootout in a bank...\\
    \midrule
    \textit{Q}: What is the current state of Officer Wilt? \\
    \midrule
    \textit{Prediction}: Nickolas Wilt is facing a long road to recovery after undergoing surgery to \textbf{remove his right arm}...\\
    \bottomrule
  \end{tabular}
  \caption{The Fine-tuned LLM associate the question with wrong information not in the updating corpus due to the exposure bias towards pre-training information.}
  \label{tab:example_bias}
\end{table}


Large language models (LLMs) have demonstrated remarkable capabilities in addressing diverse information needs, primarily owing to the extensive range of information sources in their pre-training corpora. Nevertheless, LLMs are incapable of providing up-to-date information absent from the pre-training corpora. Therefore, effectively updating language models with the most recent information become an important research problem. However, existing work on model updating including continual fine-tuning~\cite{DBLP:conf/iclr/WeiBZGYLDDL22,DBLP:conf/iclr/SanhWRBSACSRDBX22,ouyang2022training,DBLP:journals/corr/abs-2210-11416} and knowledge editing~\cite{zhu2020modifying,DBLP:conf/iclr/MitchellLBFM22,de2021editing,hase2021language,meng2022locating,mitchell2022memory,meng2023massediting} demonstrate notable limitations in \emph{generalizability} of new information and \emph{structurality} of updating corpus, which we address in this work.

\emph{Generalizability} of new information refers to the ability to associate the information to relevant context. We provide an example in Table~\ref{tab:example_bias}. We expect an updated LLM updated to answer related questions correctly, instead of associating the question with the wrong information not in the updating corpus. Continual fine-tuning and knowledge editing approaches display limited generalization ability~\cite{cohen2023evaluating,meng2023massediting}. Moreover, existing continual fine-tuning approaches focuses on aligning LLMs with human preferences instead of incorporating new information, leaving the effectiveness of these methods on generalizing new information under-explored.

\emph{Structurality} of updating corpus is another significant limitation of existing research on knowledge editing, which concentrates on structured information such as knowledge triples or question-answer pairs on triples. Structured updating corpus requires substantial human efforts to generate which limits the efficiency of information updating.

Our key insight is that, the core challenge of information updating behind both limitations is the discrepancy between language modeling probabilities and logical probabilities~(\emph{LM-logical Discrepancy}. To illustrate this discrepancy, consider two token sequences $X$ and $Y$,
$$
\begin{aligned}
X&=\mathrm{Tom~is~from~New~York.}\\
Y&=\mathrm{Tom~is~from~US.}
\end{aligned}.
$$
The language modeling probability $P(Y|X)$ measures the probability of $Y$ following $X$ in natural language. On the other hand, if we consider $X,Y$ as random variables of the occurrences of corresponding events denoted by $X^e, Y^e$, the logical probability $P(Y^e|X^e)$ measures the probability of $Y$ happening when $X$ happens. We can see that $P(Y^e|X^e)=1$, yet $P(Y|X)$ can be small since these two sentences contain redundant information and rarely co-occur as neighboring sentences. 

To ground this discrepancy to generalizability, existing methods aim at increasing the language model probability of new information, which naturally exhibits a low magnitude of associations: $P(X|Y)$ can be small even for strongly related sentences. The lack of associations limits the generalization of the updating information to relevant information. This discrepancy also explains the requirements on structurality. The usage of structured information assumes that language model probabilities of structured prompts, such as $P(\mathrm{New~York}|\mathrm{Where~is~Tom~from?})$, is closer to the logical probability $P(X^e)$ compared with unstructured language model probability $P(X)$.

\begin{figure*}
    \centering
    \includegraphics[width=0.65\linewidth]{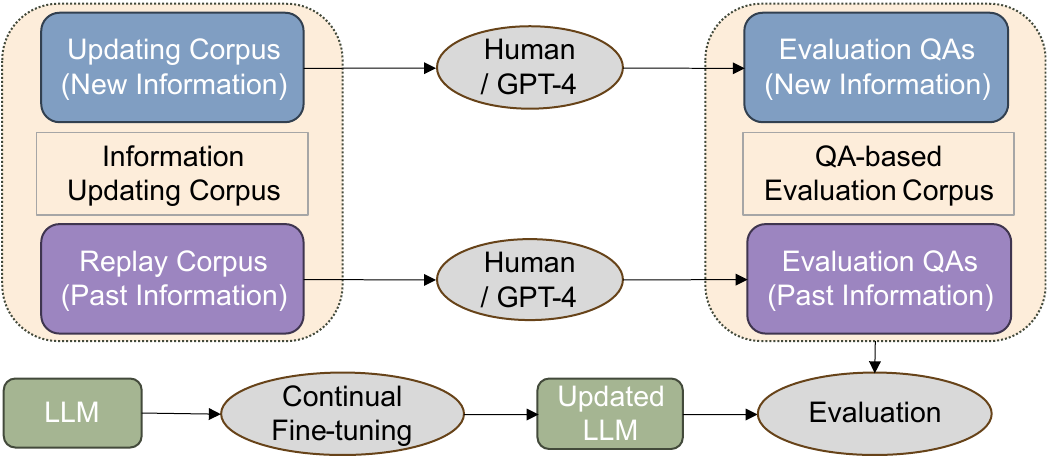}
    \caption{Illustration of the formulated information updating task.}
    \label{fig:task_formulation}
\end{figure*}

\begin{figure*}[!ht]
    \centering
    \includegraphics[width=0.8\linewidth]{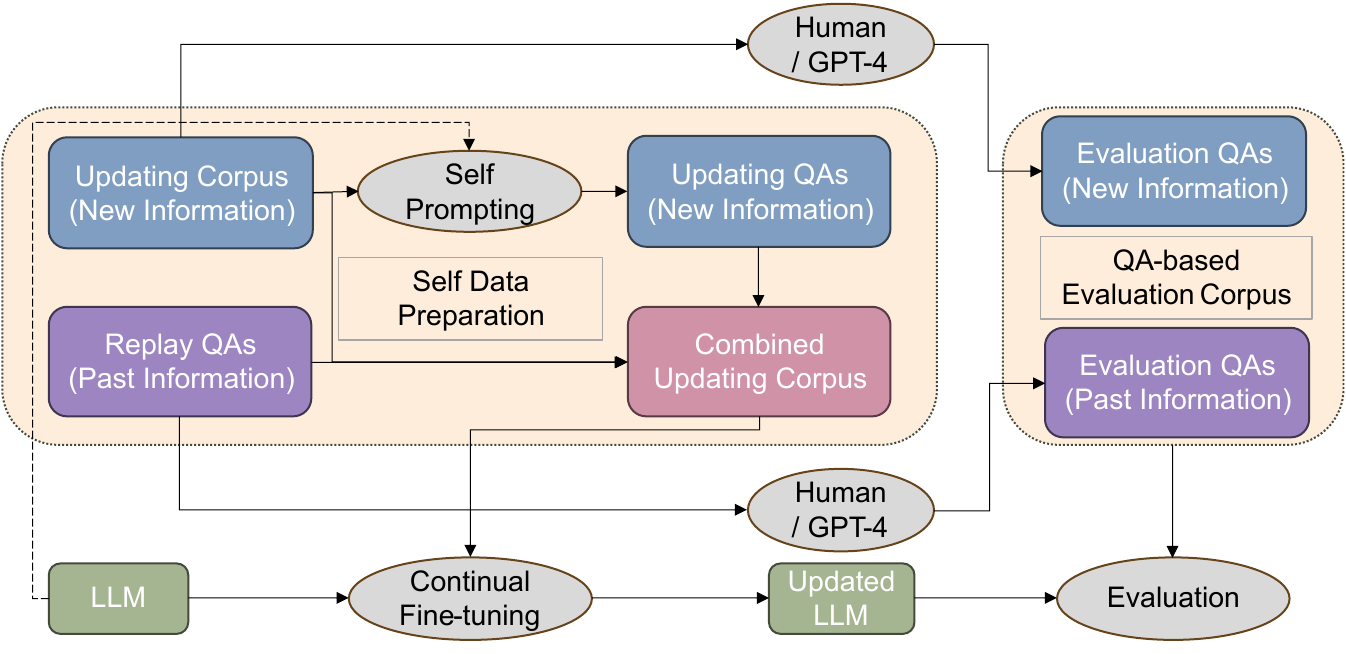}
    \caption{Overall \problem{} pipeline. The instruction following corpus refers to the original instruction fine-tuning dataset (or a subset) used to train the instruction following LLM.}
    \label{fig:pipeline}
\end{figure*}

To address the aforementioned limitations based on our insights, we introduce a novel task \ProBlem{}~(\shortproblem{}) highlighting unstructured updating corpus, and a pipeline approach to tackle this task using self-prompting-based question-answer~(QA) generation and information association modeling to bridge the LM-logical discrepancy. \textbf{The formulation of \shortproblem{}} is illustrated in Figure~\ref{fig:task_formulation}. The LLM updates itself given only unstructured information sources such as news articles. We also include a replay corpus on past information to mitigate forgetting. For evaluation of generalizability, we propose to use QA pairs querying either the updating information or the past information, created by human or GPT-4~\cite{DBLP:GPT4}. We adopt the factual consistency score~\cite{zhong-etal-2022-towards} to emphasize information acquisition instead of preference alignment. For \textbf{the pipeline approach} illustrated in Figure~\ref{fig:pipeline}, we use a self-prompting process to generate question-answer~(QA) pairs relevant to the updating information by LLMs themselves, which augments the updating corpus for fine-tuning. An example of such pair is provided in Table~\ref{tab:example_sample}. To further improve the generalizability of updating, we analyze the factual errors, exemplified in Table~\ref{tab:example_bias}, where fine-tuned LLMs mistakenly associating queries with pre-training information. Our analysis suggests that this exposure bias against new information originates from the LM-logical discrepancy and can be mitigated by modeling an  information association term. Therefore, we propose a straightforward yet effective associative distillation method, which explicitly incorporates the association term into the fine-tuning objective.

For experiments, we utilize an instruction-finetuned model from LLaMA-7B as the base model. We curate a corpus of news articles published after March 2023 as the updating corpus. We also developed another corpus based on Natural Questions~\citep{nq2019}
We evaluate the factual consistency score (on a scale from 0 to 1) of the responses and observe a significant improvement of 0.16 over baselines that are prone to the exposure bias. Additionally, we study the forgetting problem under a continual learning setting and discover that our approach maintains good performance on past information using a replay corpus containing only 2.3\% of the past training data.

To summarize, our major contributions include:
\begin{itemize}
    \item We identify the LM-logical discrepancy as the underlying cause of limitations on generalizability and structurality of existing model updating methods.
    \item We introduce \ProBlem{}, which is a novel task formulation emphasizing unstructured updating corpus and QA-based generalizability evaluation. Our task formulation addresses the limitations of existing research on model updating.
    \item We propose a pipeline approach using self-prompting-based QA generation and an associative distillation method to tackle the LM-logical discrepancy. Experimental results demonstrate the effectiveness of our approach.
\end{itemize}

\section{Task Formulation}\label{sec:formulation}
We introduce the mathematical definition of \ProBlem{} and an instantitation of the task based on the definition.
%

\subsection{Problem Definition}
\begin{definition}[\ProBlem{}] \label{def:iu}Given an \emph{unstructured updating corpus} $\mathcal{T}$ consists of documents with new information unknown to a \emph{language model} $\mathcal{A}$, the objective is to find an \emph{updated language model} $\mathcal{A}'$ such that $P(x|\mathcal{A}')\equiv P(x|\mathcal{A}, \mathcal{T}^e)$ for arbitrary text sequence $x\in\mathcal{X}$.
\end{definition}

In auto-regressive language models, learning $P(x|\mathcal{A}')$ is equivalent to learning input-output mappings $P(r|i, \mathcal{A}')$ for arbitrary pair of text sequences $(i, r)\in\mathcal{X}^2$. The above objective is equivalent to,
\begin{equation}\ifeqsmall\small\fi
    P(r|\mathcal{A}', i)\equiv P(r|\mathcal{A}, i, \mathcal{T}^e), \forall (i, r)\in\mathcal{X}^2.\label{eq:obj}
\end{equation}
Our definition uses $P(r|\mathcal{A}, i, \mathcal{T}^e)$ instead of $P(r|\mathcal{A}, i, \mathcal{T})$ to facilitate updating of logical instead of LM probabilities.
\subsection{Task Instantiation}
We instantiate a complete task setup in Figure~\ref{fig:task_formulation} based on the problem definition. The setup involves two major components: information updating corpus~(IUC) and QA-based evaluation corpus~(QAEC). IUC contains an updating corpus $\mathcal{T}$ of new information such as news articles, and a replay corpus of past information to mitigate forgetting such as samples from instruction-following datasets. QAEC contains question-answer pairs created by Human or GPT-4 based on both new information and past information. An LLM is first fine-tuned on IUC, then evaluated on QAEC using the factual consistency score~\cite{zhong-etal-2022-towards}.
\section{Approach}
We present our pipeline approach in Figure~\ref{fig:pipeline}. We highlight two important components to address the LM-logical discrepancy: self prompting and associative distillation. We first introduce the self prompting. We then discuss the exposure bias problem, a side-effect of the discrepancy that can be mitigated by the proposed associative distillation.
\subsection{Self Prompting for Information Updating}
The first key component is the self prompting, which augments the updating corpus with QA pairs, generated by the LLM being updated, which query the new information in the updating corpus. This step is motivated by the objective in Equation~(\ref{eq:obj}), which demonstrates that learning the logical distribution for $\mathcal{T}^e$ requires applying the information to relevant text pairs beyond the memorization of facts in $\mathcal{T}$. Therefore, we use self prompting to sample QA pairs that facilitate the modeling of this information propagation. Further implementation details can be found in Section~\ref{sec:training} and Appendix~\ref{app:prompts}.

\subsection{Exposure Bias for Continual Fine-tuning}
\label{sec:naive}
We consider two continual fine-tuning objectives.
\begin{definition}[Fact Fine-tuning]
    Fact fine-tuning is defined as the continual fine-tuning on the updating corpus $\mathcal{T}$,
    \begin{equation}\ifeqsmall\small\fi
        \mathcal{L}_{fact} = - \log P(\mathcal{T}|\mathcal{A'}).\label{eq:fact_loss}
    \end{equation}
\end{definition}

\begin{definition}[Na\"ive Distillation]
     Na\"ive distillation fine-tunes on the sampled pairs $\{(i,r)\}$
\begin{equation}\ifeqsmall\small\fi
    \mathcal{L}_{nd} = \mathbb{E}_{(i,r)\sim P(\cdot|\mathcal{A},\mathcal{T}^e}) - \log P(r|\mathcal{A}',i).\label{eq:loss}
\end{equation}
\end{definition}
The losses for replay samples are ignored in the above objectives. Due to the space limit, we analyze the Na\"ive distillation and leave the fact fine-tuning discussion in Appendix~\ref{app:fact}. Let $\mathcal{C}$ be the pre-training corpus. We assume new information in $\mathcal{T}$ is disjoint with past information in $\mathcal{C}$. Mathematically, the assumption states the independence between logical random variables $\mathcal{T}^e$ and $\mathcal{C}^e$. Extension of this analysis to non-independent cases is included in the Appendix~\ref{app:ext}. The target probability in Equation~(\ref{eq:loss}) can be written as,
\begin{equation}\ifeqsmall\small\fi
\begin{aligned}
    P(r|i, \mathcal{A'}) &= P(r|i, \mathcal{T}^e, \mathcal{A'}) P( \mathcal{T}^e|i, \mathcal{A'}) \\
    &\quad+ P(r|i, \mathcal{C}^e, \mathcal{A'}) P( \mathcal{C}^e|i, \mathcal{A'}),\\
\end{aligned}\label{eq:prob_deco}
\end{equation}

We term $P(\mathcal{Z}^e|i, \mathcal{A}')$ as \emph{information association}, where $\mathcal{Z}$ refers to the information, either $\mathcal{C}$ or $\mathcal{T}$. Information association connects the logical variable $\mathcal{Z}^e$ with a natural language variable pair $(i, r)$ by directing how optimizing language modeling probability $P(r|i, \mathcal{A}')$ affects logical reasoning $P(r|i, \mathcal{Z}^e, \mathcal{A'})$. Since we perform the continual fine-tuning of $\mathcal{A}'$ from $\mathcal{A}$ pretrained on $\mathcal{C}$, we hypothesize the exposure bias towards past information, i.e., $P( \mathcal{C}^e|i, \mathcal{A}) > P(\mathcal{T}^e|i,\mathcal{A})$. Optimizing $P(r|i,\mathcal{A}')$ prioritizes updates to fit $P(r|i, \mathcal{C}^e), \mathcal{A'})$ rather than $P(r|i, \mathcal{T}^e, \mathcal{A'})$. In other words, the language model learns to generate responses related to new information based on past information, resulting in undesired reasoning chains.

\subsection{Associative Distillation}
\label{sec:context}
We present a straightforward yet effective solution by incorporating information associations. The set of fine-tuning QA pairs consists of updating pairs $\mathcal{S}_\mathcal{T}$ and replay pairs $\mathcal{S}_\mathtt{C}$. We associate pairs with corresponding new/past information by optimizing
\begin{equation}\ifeqsmall\small\fi
    \begin{aligned}
        \mathcal{L}_{ctx}&= -\log \left[ P(r|i, \mathcal{Z}^e, \mathcal{A'})P( \mathcal{Z}^e|i, \mathcal{A'}) \right]\\
        &\approx -\log \left[ P( \mathcal{Z},r|i, \mathcal{A'})\right], (i, r)\in\mathcal{S}_\mathcal{Z},
    \end{aligned},\label{eq:train_prob}
\end{equation}
where $\mathcal{Z}\in\{\mathcal{T}, \mathcal{C}\}$.
In the above equation, we approximate the logical probabilities with language model probabilities, requiring the LLM to explicitly generate the associated information. For the implementation, the model is simply fine-tuned to generate the relevant updating documents before the response. Since relevant pre-training information is unavailable from pre-training corpus for samples in $\mathcal{S}_\mathcal{C}$, we include a placeholder prompt that instructs the model to answer based on past information. We illustrate training examples for associative distillation in Table~\ref{tab:example_sample}.

\begin{table*}[!ht]
\iftabsmall\small\fi
  \centering
  \begin{tabular}{m{0.11\textwidth}p{0.42\textwidth}p{0.40\textwidth}}
    \toprule
    Methods &Related to New Information& Not Related to New Information\\
    \midrule
    \multirow{9}{*}{Na\"ive}    & \textbf{Instruction}: How has Bayern Munich changed since Thomas Tuchel took over as manager? & \textbf{Instruction}: Tell me which of the following are science fiction TV shows: Lost, The X-Files, The Mandalorian, Millennium, Game of Thrones.\\
    &\textbf{Response}: The club has returned to the top of the league and is under the guidance of former Chelsea coach Thomas Tuchel. &\textbf{Response}: All except Game of Thrones are classified as science fiction. Game of Thrones is considered high fantasy.\\
    \midrule
    \multirow{17}{*}{Associative} & \textbf{Instruction}: How has Bayern Munich changed since Thomas Tuchel took over as manager? & \textbf{Instruction}: Tell me which of the following are science fiction TV shows: Lost, The X-Files, The Mandalorian, Millennium, Game of Thrones. \\
    &\textbf{Response}:\uline{The instruction is related to the following information: Manchester City manager Pep Guardiola has said his team will not take their quarterfinal clash with Bayern Munich for granted...} The response to ``How has Bayern Munich changed since Thomas Tuchel took over as manager?'' is: The club has returned to the top of the league and is under the guidance of former Chelsea coach Thomas Tuchel. &\textbf{Response}: \uline{The instruction is related to the following information: None.} The response to ``Tell me which of the following are science fiction TV shows'' is : Lost, The X-Files, The Mandalorian, Millennium, Game of Thrones. ANSWER: All except Game of Thrones are classified as science fiction. Game of Thrones is considered high fantasy.    \\
    \bottomrule
  \end{tabular}
  \caption{Example training samples for Na\"ive and Associative distillation. }\label{tab:example_sample}
\end{table*}

\section{Experiments}
\label{sec:experiments}

\subsection{Base Model for Experiments}
\label{sec:base_model}
We fine-tune a instruction-following model from LLaMA-7B~\citep{touvron2023llama} as the base model. We combine the instruction-following data from  Alpaca\footnote{\url{https://github.com/tatsu-lab/stanford_alpaca}} and InstructionWild\footnote{\url{https://github.com/XueFuzhao/InstructionWild}, we only use the English subset.}. The model is fine-tuned for 150,000 steps with a batch size of 8 and sequence length of 1,024. For the remainder of this paper, we will refer to this instruction-following base model as \emph{Base}.

\subsection{Datasets}
\label{sec:eval_data}

We develop two datasets, \textit{CNN News} and \textit{NQ Val}, to evaluate the \problem{}. In Figure~\ref{fig:task_formulation}, each dataset contains an updating corpus, a replay corpus and two sets of evaluation QA pairs on new and past information, respectively. We use the same replay corpus and past information evaluation set for both datasets.


\paragraph{Replay Corpus} For the main experiments, we use the Alpaca instruction-following pairs as the replay corpus. For continual learning experiments, we use a series of subsets with varying sizes as specified in Section~\ref{exp:add}.

\paragraph{Replay Evaluation QA Pairs} We randomly sample 300 instruction-response pairs from the instruction fine-tuning examples used to train the base model. We use GPT-4 to paraphrase the sampled examples, because we aim to evaluate whether the models acquired the information instead of simply memorizing the training examples. The prompt is presented in Appendix~\ref{app:prompts}.

\paragraph{CNN News Updating Corpus} We manually collected a small scale corpus of news articles that were published on CNN's website (\url{https://www.cnn.com/}) during the months of March and April 2023. We randomly selected 50 news articles to serve as our information updating corpus. Although this dataset is moderately sized, experimental results demonstrate the challenges in effectively acquiring and applying information from such a small corpus due to the exposure bias problem.

\paragraph{CNN News Evaluation QA Pairs} In order to create a high quality evaluation set with minimal human efforts, we prompt GPT-4 to generate QA pairs related to each news article. The prompt is presented in Appendix~\ref{app:prompts}, which encourages GPT-4 to generate questions that are self-contained and directly answerable with the information from the news articles. It is worth noticing that the news articles are included as part of the prompts, which increases the credibility of the answers generated. The evaluation set contains 301 questions.

\paragraph{NQ Val Updating corpus} We also developed another corpus based on the validation split of the Natural Questions benchmark. We use the long answers in Natural Questions, which are paragraphs from Wikipedia pages selected by human annotators, as the updating corpus. Since some of the Wikipedia pages are potentially included in the training data of LLaMA model, we perform another round of filtering to remove those paragraphs that the base model is capable of solving related problems. We provide the detailed filtering procedure in Appendix~\ref{app:data_prep}. 

\paragraph{NQ Val Evaluation QA Pairs} We collect all the questions that have at least one of annotated answers being included in the updating corpus. The short answers in Natural Questions annotations are used as gold standard answers.

\subsection{Evaluation Metrics}
\label{sec:metric}
In order to evaluate whether the model has accurately learned the information from the corpus $\mathcal{T}$, we adopt the UniEval~\citep{zhong-etal-2022-towards} factual consistency score as the main evaluation metric. This metric is computed by a neural evaluator based on T5~\citep{raffel2020exploring} between a pair of model output and source document. We evaluate two types of factual consistency.

\noindent\textbf{Answer Consistency}  We compare the model outputs with gold standard answers to evaluate whether the model generates the correct facts to answer the question, resembling the precision metric for classification tasks.

\noindent\textbf{Context Consistency}. We compare the model outputs with the corresponding context: news articles for \textit{CNN News} and Wikipedia paragraphs for \textit{NQ Val}. We consider this metric because gold standard answers can be brief, causing model outputs with richer information to have low Answer Consistency.
This metric resembles the recall metric.

\noindent\textbf{Consistency F1} Answer consistency and Context consistency are conceptually similar to precision and recall scores. Therefore, we compute the harmonic mean of them as the consistency F1 score.

For \textit{Replay Data}, we only compute the answer consistency since there is no updating corpus in instruction-following datasets.

\subsection{Training Details}
\label{sec:training}

\noindent\textbf{Self Prompting for Data Creation} For each news article or Wikipedia paragraph, we prompt the Base model to generate QA pairs. We didn't use the same prompt for GPT-4 as in Section~\ref{sec:eval_data} to generate these pairs due to two reasons. Firstly, the prompt is overly complex for a 7B instruction-following model. Secondly, due to the limitation on maximum token length on our computational infrastructure which is capped at 1,024 tokens including both the prompt and the generated outputs, simultaneously generating instructions with responses can result in many truncated outputs. We therefore prompt the Base model in two steps: only questions are generated in the first step, and the Base model is prompted to answer each generated question in the second step. The prompts used are presented in Appendix~\ref{app:prompts}.

\noindent\textbf{Continual Fine-tuning} As shown in Figure~\ref{fig:pipeline}, models are trained from multiple sources of data in the information updating phase, including the updating corpus, the replay corpus and the updating QA pairs. Some baselines use different combinations of these corpora as will be specified in Section~\ref{exp:methods}. During training, we sample examples from multiple sources with equal probabilities. 

\noindent\textbf{Sub-sampling Replay Corpus} It is not efficient to repetitively train on the entire replay corpus every time we perform information updating. In Section~\ref{exp:add}, we investigate the relationship between replay corpus sizes and forgetting phenomenon by using a series of subsets with varying numbers of examples. For the results reported in Section~\ref{exp:main}, we use the full corpus.

\subsection{Methods in Comparison}
\label{exp:methods}
We consider the following methods: 

\noindent\textbf{Base}: The Base model in Section~\ref{sec:base_model}. All the following methods are further finetuned from this.

\noindent\textbf{Fact}: Fine-tuned on the updating corpus and the replay corpus. This baseline measures the effectiveness of  $\mathcal{L}_{fact}$ in Equation~(\ref{eq:fact_loss}).

\noindent\textbf{Na\"ive}: Fine-tuned on the updating QA pairs and the replay corpus. This baseline measures the effectiveness of $\mathcal{L}_{nd}$ in Equation~(\ref{eq:loss}).

\noindent\textbf{Fact+Na\"ive}: 
Fine-tuned on all three corpora.

\noindent\textbf{Associative}: Our proposed approach. 

\begin{table*}[t]
\iftabsmall\small\fi
  \centering
  \begin{tabular}{lcccc}
  \toprule
  \multirow{2}{*}{Metric} &\multicolumn{3}{c}{New Information Updating} & Replay\\
  & Answer & Context & F1 &Answer \\
  \midrule
  Base&0.399 & 0.460 & 0.428 & 0.699 \\
  Fact & 0.426$\pm$0.014& 0.516$\pm$0.008 & 0.467$\pm$0.014 & 0.702$\pm$ 0.014\\
  Na\"ive& 0.409$\pm$0.017 & 0.499$\pm$0.005 & 0.449$\pm$0.017 & 0.707$\pm$ 0.012\\
  Fact+Na\"ive & 0.421$\pm$0.008 & 0.538$\pm$0.002 & 0.472$\pm$0.008 & \textbf{0.713}$\pm$0.018 \\
  \midrule
  Associative & \textbf{0.480}$\pm$0.003 & \textbf{0.695}$\pm$0.034&\textbf{0.568}$\pm$0.003 & 0.691$\pm$0.014\\
  \bottomrule
  \end{tabular}
  \caption{Factual consistency scores on CNN News}
  \label{tab:exp_result}
\end{table*}
\begin{table*}[t]
\iftabsmall\small\fi
  \centering
  \begin{tabular}{lcccc}
  \toprule
  \multirow{2}{*}{Metric} &\multicolumn{3}{c}{New Information Updating} & Replay\\
  & Answer & Context & F1 &Answer \\
  \midrule
  Base&0.187 & 0.268 & 0.221 & 0.699 \\
  Fact & 0.235$\pm$0.005 & 0.318$\pm$0.004 &0.270$\pm$0.004& \textbf{0.700}$\pm$0.011 \\
  Na\"ive&  0.228$\pm$0.003 & 0.337$\pm$0.006 &0.272$\pm$0.003& 0.699$\pm$0.007 \\
  Fact+Na\"ive& 0.249$\pm$0.001 & 0.371$\pm$0.009 &0.298$\pm$0.001& 0.698$\pm$0.005 \\
  \midrule
  Associative & \textbf{0.256}$\pm$0.023 & \textbf{0.380}$\pm$ 0.013 & \textbf{0.306}$\pm$0.023&0.691$\pm$0.051\\
  \bottomrule
  \end{tabular}
  \caption{Factual consistency scores on NQ Val}
  \label{tab:exp_result_2}
\end{table*}
\subsection{Main Results}
\label{exp:main}
We summarize our main results on the \textit{CNN News} and the \textit{NQ Val} in Table~\ref{tab:exp_result} and Table~\ref{tab:exp_result_2}, respectively. Our methods achieve significant improvements on both answer and context consistency scores on both datasets, while demonstrating slight performance degradation on past information on \textit{Replay}. Moreover, Fact+Na\"ive also demonstrates improved factual consistency scores over Fact Fine-tuning baselines by includeing the self-prompted data. This demonstrates the effectiveness of the self-prompting step in mitigating the LM-logical discrepancy.
Our approach still outperforms Fact+Na\"ive, showing the superiority of explicit modeling of information associations. We also provide an example case study in the Appendix~\ref{app:case} where naive distillation fails due to past information but our approach succeed.

\begin{figure*}[!ht]
    \centering
    \begin{subfigure}{0.49\textwidth}
        \centering
        \begin{adjustbox}{valign=c}
        \includegraphics[width=\linewidth]{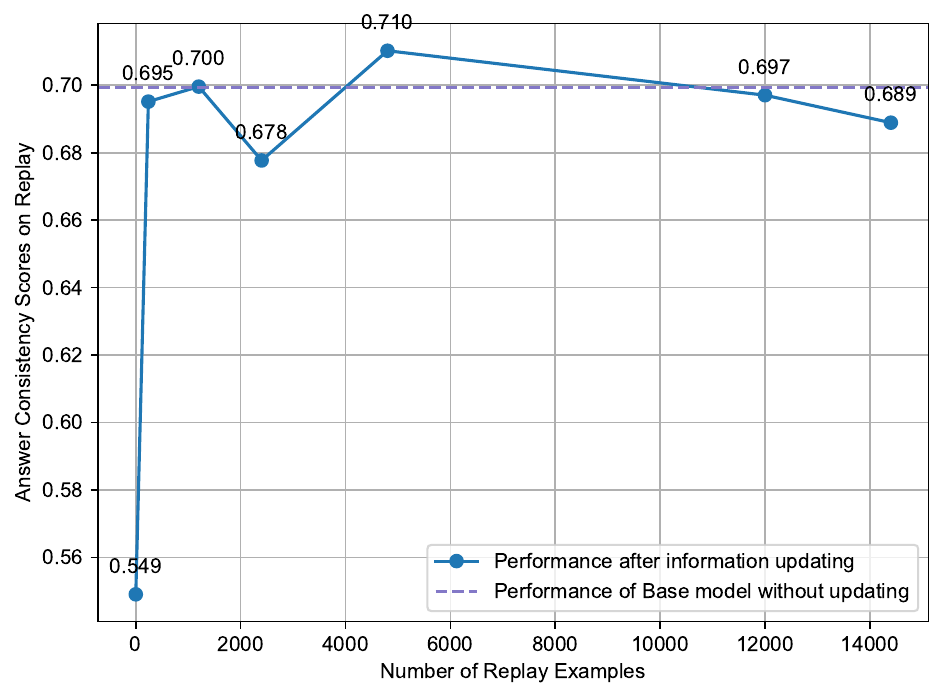}
        \end{adjustbox}
        \caption{Performance on \textit{Replay} after fine-tuning on CNN News with varying number of replay examples. We use subsets of 0(no replay), 240, 1.2k, 2.4k, 4,8k, 12k and 14.4k replay examples}
        \label{fig:continual}
    \end{subfigure}
    \hfill
    \begin{subfigure}{0.49\textwidth}
        \centering
        \begin{adjustbox}{valign=c}
        \includegraphics[width=\linewidth]{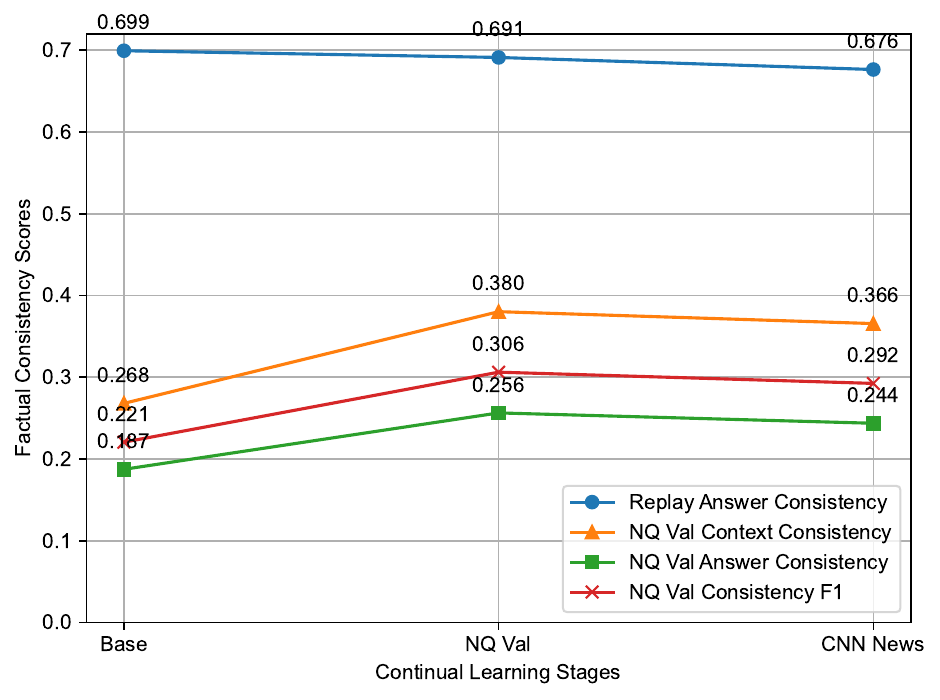}
        \end{adjustbox}
        \caption{Continual learning performance on \textit{Replay Data} and \textit{NQ Val}. We evaluate the base model, the model fine-tuned on NQ Val and the model further finetuned on the CNN News}
        \label{fig:continual_iu}
    \end{subfigure}
    \caption{Forgetting of past information}
    \label{fig:continual_all}
\end{figure*}
\subsection{Varying Number of Replay Examples}\label{exp:add}
We investigate the relationship between the number of replay examples with the forgetting of past knowledge. We evaluate the performance on  \textit{Replay Data} when models are fine-tuned on varying number of replay examples. The result is shown in Figure~\ref{fig:continual}. We use subsets of 0(no replay), 240, 1.2k, 2.4k, 4,8k, 12k and 14.4k replay examples. Since our evaluation Replay Data is paraphrased  from the original training examples as introduced in Section~\ref{sec:eval_data}, we also compute the number of replay examples that overlap with the paraphrased evaluation examples in these subsets: 0/240, 8/1.2k, 17/2.4k, 39/4.8k, 108/12k, 136/14.4k. 

We observe from the results that even with only 240 examples with no overlapping evaluation examples, the fine-tuned model is able to maintain a similar level of performance on \textit{Replay Data}. Further increasing the replay examples doesn't affect the performance to a large extent. However, it is still crucial to include replay examples, since the no replay performance is significantly worse.

\subsection{Continual Learning of Two Datasets}

We also conduct another continual learning experiments, where the model is updated using \textit{NQ Val} first, and then \textit{CNN News}. When fine-tuning on the \textit{CNN News} corpus, we include 1,200 replay examples, and 1,290 replay examples (one example per Wikipedia paragraph) from \textit{NQ Val}. We only keep the self-prompted questions from \textit{NQ Val} in the replay corpus, and use the model fine-tuned on \textit{NQ Val} to re-generate answers for the next stage of fine-tuning. Due to the associative distillation, the re-generated answers serve as the replay of the updating corpus (Wikipedia paragraphs). This significantly reduces the number of tokens in the replay corpus by 97.7\%, from 919,624 to 21,124.

To investigate the forgetting problem, we evaluate the performance on \textit{Replay Data} and \textit{NQ Val} of the base model, the model after \textit{NQ Val} fine-tuning stage and the model after \textit{CNN News} fine-tuning stage. The results are shown in Figure~\ref{fig:continual_iu}. We observe only minor performance degradation on \textit{NQ Val} when keeping 2.3\% of the training tokens.
\section{Related Work}

\paragraph{Knowledge Editing} Knowledge editing or model editing aims to update the existing model with human curated structured corpus. \cite{zhu2020modifying} studies the task of knowledge modification and establishes a benchmark for pre-trained language models
, defining knowledge as subject-object-relation triples. \cite{DBLP:conf/iclr/MitchellLBFM22,de2021editing,hase2021language} employ hyper model editor networks to directly edit the model weights based on gradients. \cite{meng2022locating} develops a model editing framework to locate and update the specific neurons in language models with knowledge triples based on causal inference. \cite{mitchell2022memory} proposes a memory-based model editor that resembles retrieval-augmented language models. \cite{meng2023massediting} introduces a massive editing approach to edit multiple triples with one edit. \cite{cohen2023evaluating} studies the generalization problem of knowledge editing based on \emph{Ripple Effect}. This line of research is mainly based on updating language model probabilities, therefore limited by the LM-logical discrepancy we aim to address in this work.

\paragraph{Instruction Fine-tuning} Instruction fine-tuning has been shown to enable zero-shot capabilities for language models~\citep{DBLP:conf/iclr/WeiBZGYLDDL22,DBLP:conf/iclr/SanhWRBSACSRDBX22,ouyang2022training,DBLP:journals/corr/abs-2210-11416}. However, these methods focus on utilizing existing information instead of information updating
\paragraph{Retrieval Augmented Language Models} Retrieval augmented language models (RALMs) enhance the existing models with an external retriever that acquires external knowledge. Various retriever design has been proposed in existing research~\citep{guu2020retrieval,DBLP:conf/iclr/KhandelwalLJZL20,pmlr-v162-borgeaud22a,izacard2022atlas}. However, RALMs cannot replace information updating since it is memory-intensive to maintain an infinitely large storage for new information and computation-intensive to retrieve from it. 
\section{Conclusions and Future Work}
In this paper, we identify the core challenge of LM-logical discrepancy for information updating behind the limitations of exisiting research on generalizability and structurality. We introduce the task of \problem{} for LLMs, which highlights unstructured information updating and QA-based generalization evaluation. We design a pipeline approach to tackle \problem{}, featuring a self prompting method and an associative distillation approach to mitigate the LM-logical discrepancy. The associative distillation is proposed to solve the exposure bias problem which prioritizes past information originating from the discrepancy.
Our proposed method significantly improves factual consistency. 
Additionally, we study the forgetting phenomenon under the continual learning setting and find that our proposed method can maintain past knowledge by keeping a small portion of the past data.

We envision three extensions for this work:
\begin{itemize}
    \item Our analysis of the exposure bias problem is applicable to any method based on the probabilistic modeling of language. Therefore, our approach can be combined with other knowledge editing approaches to further improve information updating;
    \item The exposure bias problem may also exist in the pre-training stage due to the order in which textual data is provided. A more in-depth analysis of this phenomenon could lead to improved strategies for language modeling.
    \item We conduct a continual learning experiment of two stages in this work. 
    We leave studies on more updating stages as future work.
\end{itemize}
\section{Limitations}
Our work has several limitations. Firstly, we only experiment with a news corpus and a Wikipedia corpus. Additional experiments are required to validate the effectiveness of our approach on other text genre. Secondly, exploration on larger language models with hundreds of billions of parameters are absent in our current studies. Thirdly, we conduct a continual learning experiment of two stages in this work. Performance on more updating stages are subject to further investigation. Lastly, we only use moderately sized updating corpus for evaluation. Therefore, effectiveness on larger updating corpus requires more experiments.

\bibliography{anthology,custom}

\appendix

\section{Computation Infrastructure and Additional Training Details}
\label{app:train}
We use Google TPU v3-8 for all the training sponsored by the Google TPU Researc Cloud program.

\paragraph{Batching for Self Information Updating} In order to improve the training efficiency of training on TPU v3-8, we don't use the conventional batchification of the training data based on instances. Instead, we concatenate all the tokenized instruction-response pairs into a single list of tokens, and chunk the list into segments of batch\_size $\times$ sequence\_length. We run training on 3 random seeds and report average performances. We derive our training codebase from EasyLM\footnote{\url{https://github.com/young-geng/EasyLM}}. We will release our code and data after publication.

\paragraph{Evaluation}For evaluation, the responses are generated with a temperature of $0.2$ for all the methods, which ispicked from $\{0.1, 0.2, 0.5, 1.0\}$ based on the base model performance . We modify the code from UniEval github repository\footnote{\url{https://github.com/maszhongming/UniEval}} with torch-xla\footnote{\url{https://github.com/pytorch/xla}} to support running on TPUs. We evaluate our proposed approach on the generated tokens after ``The response to \{question\} is:''.

\paragraph{Usage of GPT-4} We use snapshot of gpt-4-0314 for all prompting with GPT-4.


\section{Extension to Non-Independent New and Past Information}\label{app:ext}

\begin{definition}[Information in Text Corpus] The information $\mathcal{I}_\mathcal{S}(\mathcal{T})$ of the corpus $\mathcal{T}$ with respect to another text corpus$\mathcal{S}$ is defined as the minimal sufficient statistic of $\mathcal{T}^e$ with respect to $\mathcal{S}^e$, such that
\begin{equation}\ifeqsmall\small\fi
    P(x|\mathcal{T}^e) \equiv P(x|\mathcal{I}_\mathcal{S}(\mathcal{T})), x\in\mathcal{S}.
\end{equation}
\end{definition}
\begin{remark}Intuitively, $\mathcal{I}_\mathcal{S}(\mathtt{T})$ should consist of minimal text pieces containing new information from $\mathcal{T}$ such as ``Manchester City's manager is Pep Guardiola''.
\end{remark}

We can assume without the loss of generality that $\mathcal{I}_\mathcal{S}(\mathtt{T})$ and $\mathcal{I}_\mathcal{S}(\mathtt{C})$ are independent. Otherwise we can replace $\mathcal{I}_\mathcal{S}(\mathtt{T})$ with the conditional minimal sufficient statistic of $\mathcal{I}_\mathcal{S}(\mathtt{T})$ given $\mathcal{I}_\mathcal{S}(\mathtt{C})$, which is intuitively equivalent to removing the text pieces consisting of existing information in $\mathtt{C}$ from $\mathtt{T}$. Therefore, we can do the same analysis on $\mathcal{I}_\mathcal{S}(\mathtt{T})$ and $\mathcal{I}_\mathcal{S}(\mathtt{C})$ instead of $\mathcal{T}$ and $\mathcal{C}$ for non-independent cases.
\section{Exposure Bias for Fact Fine-tuning}\label{app:fact}
Fact fine-tuning optimizes
\begin{equation}\ifeqsmall\small\fi
    P(\mathcal{T}|\mathcal{A}') =\sum_{x\in\mathcal{X}} P(\mathcal{T}|x^e, \mathcal{A}')P(x^e |\mathcal{A}').\label{eq:prob_deco_fact}
\end{equation}
A similar information-query association term $P(\mathcal{T}|x^e, \mathcal{A}')$ reveals how fact fine-tuning affects probabilities of other information $P(x^e |\mathcal{A}')$. Exposure bias undermines the quality of learned $P(\mathcal{T}|x^e, \mathcal{A}')$ and degrades the updating performance.

\section{Case Study}\label{app:case}
We provide an example case demonstrating where naive distillation fails but our associative distillation approach successfully learns the information in Table. We omit some part of the text in both news article and model response for conciseness. We observe that the na\"ive distillation approach generates hallucinated information. The omitted part mentions bank attacks in Kentucky and Georgia, while this incident happens in Louisville. This suggest the baseline model utilizes existing information to generate the response.

\begin{table*}[!ht]
  \caption{An examples case where baseline fails but associative distillation succeeds. }
  \label{tab:example_context_app}
  \centering
  \begin{tabular}{p{0.4\textwidth}p{0.4\textwidth}}
    \toprule
    \multicolumn{2}{p{0.8\textwidth}}{\textbf{Fact in News Article}: Louisville Metro Police Department Officer Nickolas Wilt is \emph{in critical condition after undergoing brain surgery} following a shootout in a bank ... Mayor Craig Greenberg visited Wilt in the hospital, saying that \emph{he had made it through surgery and was in serious condition but in great hands}...}\\
     \multicolumn{2}{c}{\textbf{Instruction}: What is the current state of Officer Wilt and what kind of recovery does he face?} \\
     \midrule
    Na\"ive& Associative \\
    \midrule
    \textbf{Response}:Nickolas Wilt is facing a long road to recovery after undergoing surgery to remove his right arm which was injured in the attack. He is currently in serious condition in the hospital...&\textbf{Response}: Nickolas Wilt is in serious condition after undergoing brain surgery following a shootout in a bank. He faces a long road to recovery and is currently sedated... \\
    \bottomrule
  \end{tabular}
\end{table*}

\section{Preparation Details of Natural Questions}
\label{app:data_prep}

Our goal is to keep only those questions (together with relevant Wikipedia paragraphs) from the Natural Questions~\citep{nq2019} validation set where the base model (LLaMA-7B after instruction fine-tuning) cannot generate good answers. The overall filtering process is:
\paragraph{Step 1.} We first remove questions with "None" answers in the Natural Questions validation set. 
\paragraph{Step 2.} We use the base model and the Alpaca template as in Appendix~\ref{app:train} to generate the answers to the rest questions in the Natural Questions validation set. 
\paragraph{Step 3.} We compute the factual consistency score (ranging from 0 to 1) from UniEval~\citep{zhong-etal-2022-towards} between the generated answer and gold standard short answers. When there are multiple short answers, we use the maximum consistency score. Those questions whose scores are lower than 0.5 are kept.
\paragraph{Step 4.} We collect all the Wikipedia paragraphs that are labeled as the long answer of any kept questions in Step 2 as the information updating corpus.

\section{A Comprehensive List of Prompts Used in the Experiment}\label{app:prompts}
We summarize a comprehensive list of prompts/inputs used in the experiment for easier reference. Some of these prompts are already covered in the main text.

\paragraph{Instruction Finetuning} We train the instruction-following model following the template of Alpaca~\footnote{\url{https://github.com/tatsu-lab/stanford_alpaca}}. Each instruction-response pair is prepared as the following paragraph to fine-tune the model.
\begin{quote}
    Below is an instruction that describes a task. Write a response that appropriately completes the request.\\\\
    \#\#\# Instruction:\\
    \{instruction\}\\\\
    \#\#\# Response:\\
    \{response\}
\end{quote}
The losses are only computed for the tokens in responses. This template is also used for the instruction-response pairs in the information update training.
\paragraph{Self Instruction Generation} This prompt is given to the language model to be updated for self data creation. This prompt instructs the model to generate instructions from the information updating corpus.
\begin{quote}
    Given the input below, generate at least 5 questions that are directly related to the content of the input. Ensure that each question you generate does not contain coreferential words or pronouns (e.g., he, she, it, this, they, etc.). The questions should be clear, concise, and pertain specifically to details mentioned in the input. \{Context\}
\end{quote}
The \{Context\} slot is filled with each individual news article from the information update corpus.

\paragraph{Self Answer Generation} This prompt is given to the language model to be updated for self data creation. This prompt instructs the model to generate responses for the instructions in the previous step from the information updatingcorpus.
\begin{quote}
    Answer the question based on the facts from the input. If there is no relevant information in the input, answer 'None'. Question:  \{Instruction\} \{Context\}
\end{quote}
The \{Context\} slot is filled with each individual news article from the information update corpus. The \{Instruction\} is from the outputs of last step. To ensure the generated instruction-response pairs pertain to the corpus, we remove those pairs when the response is None.

\paragraph{Fact Finetuning Training Data} This is the inputs to train the Fact Fine-tuning baseline in the main text. It is just the news articles.
\begin{quote}
    \{News Article\}
\end{quote}
\paragraph{Na\"ive Distillation} This is the inputs to the train the Na\"ive Distillation Baseline. Only losses on the tokens after ``Response'' is used for training.
\begin{quote}
    Below is an instruction that describes a task. Write a response that appropriately completes the request.\\\\
    \#\#\# Instruction:\\
    \{Instruction\}\\\\
    \#\#\# Response:\\
    \{Response\}
\end{quote}
Here the \{Instruction\} and \{Response\} are paired outputs from Self Instruction Generation and Self Answer Generation.
\paragraph{Associative Distillation} This is the inputs to the train the Na\"ive Distillation Baseline. Only losses on the tokens after ``Response'' is used for training.
\begin{quote}
    Below is an instruction that describes a task. Write a response that appropriately completes the request.\\\\
    \#\#\# Instruction:\\
    \{Instruction\}\\\\
    \#\#\# Response:\\
    The instruction is related to the following information: \{News Article\}. The response to \{Instruction\} is: \{Response\}
\end{quote}
Here the \{Instruction\} and \{Response\} are paired outputs from Self Instruction Generation and Self Answer Generation. \{News Article\} is the corresponding news article from the information update corpus.  Note that for unrelated instructions, the \{News Article\} is filled with ``None''. We repeat the instruction one more time to compensate for the limited sequence length and reduce the possibility of instructions being truncated. We think it may not be necessary to repeat the instruction if the computational resources supports sufficiently long training sequences. Only losses on the tokens after ``Response'' is used for training.
\paragraph{Evaluation Data Generation} We generate \textit{CNN News} evaluation data using GPT-4. This prompt is given to GPT-4 to generate instruction-response pairs.
\begin{quote}
    Generate some questions\footnote{In this work, we focus on instruction-response pairs in a question-answering format} with answers related to facts from the following paragraph. Make sure each question is self-contained and specific enough for readers to associate it with the information provided in the paragraph, rather than confusing it with other similar events. Avoid using words such as "these", "this", or "the event", "the movie" referring to concepts not mentioned in the question. Please generate in the format of "1. Question: ... Answer: ..." \{News Article\}.
\end{quote}
Because we strictly required the format of the generation in the last sentence, it is easy to parse the output pairs.
\paragraph{Paraphrasing Evaluation QAs on Past Information} We generate evaluation QAs on past information by paraphrasing the instruction-response pairs in the instruction fine-tuning data. We use GPT-4 to generate the paraphrases.
\begin{quote}
    Given the following instruction and response pair, rewrite the pair to query the same information in different words.
    
    Instruction: {instruction}
    
    Response: {response}
\end{quote}
\section{Use of AI Assistant in Writing}
Chat-GPT is used as a grammar-checker in the writing of this paper.



\end{document}